%% file: paper.tex
\documentclass[letterpaper, 10 pt, conference]{ieeeconf}
\IEEEoverridecommandlockouts    
\input{stachnisslab-latex}


\usepackage{flushend}

\title{\LARGE \bf Uncertainty-Informed Active Perception \\ for Open Vocabulary Object Goal Navigation}

\author{Utkarsh Bajpai \and Julius R\"uckin \and Cyrill Stachniss \and Marija Popovi\'{c}
  \thanks{Utkarsh Bajpai, Julius R\"uckin, Cyrill Stachniss are with the Center for Robotics, University of Bonn, Germany. Marija Popovi\'{c} is with the MAVLab, TU Delft, Netherlands. Cyrill Stachniss is also with the Lamarr Institute for Machine
Learning and Artificial Intelligence, Germany.}%
  \thanks{This work has been funded 
  by the Deutsche Forschungsgemeinschaft (DFG, German Research Foundation) under Germany's Excellence Strategy, EXC-2070 -- 390732324 -- PhenoRob and by the German Federal Ministry of Education and Research (BMBF) in the project ``Robotics Institute Germany", grant No. 16ME0999.
  Corresponding author: utkarsh@uni-bonn.de.
  }%
}

\begin{document}
\thispagestyle{empty}
\pagestyle{empty}
\maketitle

\begin{abstract}
  %



Mobile robots exploring indoor environments increasingly rely on vision-language models to perceive high-level semantic cues in camera images, such as object categories. Such models offer the potential to advance robot behaviour for tasks such as object-goal navigation~(ObjectNav), where the robot must locate objects specified in natural language by exploring the environment. Current ObjectNav methods focus on prompt engineering for perception and do not address the semantic uncertainty induced by variations in prompt phrasing. Ignoring semantic uncertainty can lead to suboptimal exploration, which in turn limits performance. Hence, we propose a semantic uncertainty-informed active perception pipeline for ObjectNav in indoor environments. We introduce a novel probabilistic sensor model for quantifying semantic uncertainty in vision-language models and incorporate it into a probabilistic geometric-semantic map to enhance spatial understanding. Based on this map, we develop a frontier exploration planner with an uncertainty-informed multi-armed bandit objective to guide efficient object search. Experimental results demonstrate that our method achieves ObjectNav success rates comparable to those of state-of-the-art approaches, without requiring extensive prompt engineering.

  
\end{abstract}

\section{Introduction}
\label{sec:intro}

Robots benefit from effective goal-directed exploration in human-centric environments to accomplish tasks such as locating and delivering objects, as well as assisting with household activities. These tasks require a robot to detect objects and understand spatial layouts with task-relevant contextual semantic cues. To this end, grounding language instructions in perceptual and spatial representations of the scene, also known as open-vocabulary perception, is key to efficient exploration and success of the task. However, open-vocabulary perception is inherently uncertain due to the wide variety of objects that exist in households and the many ways they can be described in natural language. Quantifying this perception uncertainty and accounting for it during exploration is an important element to realise reliable task execution.

This work examines the object goal navigation~(ObjectNav) task~\cite{batra2020objectnav}, in which a robot must locate and navigate to a target object in an unfamiliar indoor environment within a limited mission time. The robot needs to recognise arbitrary and diverse objects based on a natural language description. The environment is unknown prior to a mission and is only partially observable due to noisy RGB and depth sensor observations, limited in range. To this end, the robot must continuously refine its exploration strategy online using its current sensor observations to navigate to the target object.
\begin{figure}[t]
  \centering
  \includegraphics[width=0.95\linewidth]{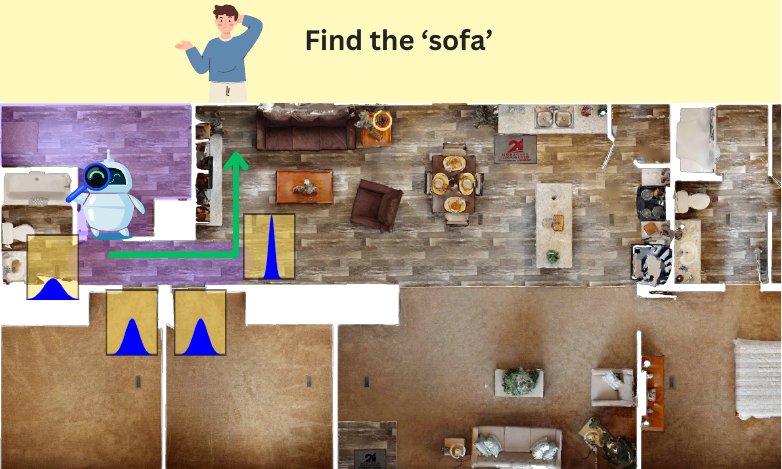}
  \caption{We develop an uncertainty-informed, open-vocabulary ObjectNav pipeline for locating arbitrary objects in indoor environments. The figure visualises our approach: given a target object, the robot needs to navigate to it (green arrow) in an initially unknown environment. The robot actively selects a frontier to explore at each timestep amongst all available frontiers (yellow rectangles) using our multi-arm bandit frontier planner informed by semantic relevance estimates about each frontier (blue Gaussians) from our probabilistic geometric-semantic map (purple).}
  \label{fig:motivation}
\end{figure}

A common strategy employed by ObjectNav methods is to construct a 2D map of the environment online using observed geometric information, such as spatial layout and obstacles. Geometric maps enable tracking explored areas and guiding exploration toward map corners~\cite{luo2022iros} or frontiers between known and unknown space, known as frontier-based exploration~\cite{yamauchi1997cira}. Semantic information, such as object categories, can also be integrated into maps using deep learning-based semantic segmentation~\cite{chaplot2020nips}, which enables exploration of areas with semantic similarity to the target object. However, most approaches are closed-vocabulary, i.e. limited to a fixed set of objects defined at training time and do not scale well to the wide variety of objects found in the real world. State-of-the-art approaches for ObjectNav~\cite{yokoyama2024vlfm,chen2023rss,gadre2023cow,majumdar2022zson} leverage open-vocabulary perception using vision-language models~(VLMs)~\cite{radford2021learning,li2023blip2}, allowing for semantically informed exploration scalable to potentially infinite object categories. Trained on large-scale image-text pairs, VLMs provide semantic relevance between arbitrary user-specified text prompts and images. 
However, VLMs are often sensitive to prompt phrasing. Variations in prompts, such as ``chair,” ``a chair,” or ``there is a chair nearby” produce different semantic relevance scores from VLMs, given the same image. This can lead to inconsistency in navigating to the target object in the ObjectNav scenario. Existing approaches~\cite{chen2023rss, yokoyama2024vlfm} often rely on fixed hand-tuned prompts appended to object names, which ignore this inconsistency and do not address this semantic uncertainty in VLM predictions. Ignoring semantic uncertainties is suboptimal and does not exploit the full potential of planning algorithms. Our paper aims to overcome this within the ObjectNav problem.



The main contribution of this paper is a semantic uncertainty-informed active perception framework for ObjectNav illustrated in \figref{fig:motivation}. We propose a probabilistic geometric-semantic mapping method that updates geometric and semantic environment information online, while explicitly modelling the uncertainty in VLM-predicted semantic relevance. We propose a probabilistic sensor model using prompt ensembling~\cite{lester2021promptensemble}, i.e. querying the VLM with varied prompts to approximate uncertainty in semantic relevance. Based on our probabilistic geometric-semantic map, we introduce a novel frontier planner with a multi-armed bandit objective for efficient exploration in ObjectNav. Our planner balances exploration of uncertain regions and exploitation of regions likely to contain the target object. By accounting for uncertainty in perception and planning, our framework eliminates the need for hand-tuned prompts.

In sum, we make the following claims. First, the success of prior open-vocabulary ObjectNav approaches relies on extensive prompt engineering, with the choice of prompts impacting downstream navigation performance.
Second, we show that semantic relevance scores from a VLM prompt-ensemble are approximately normally distributed to validate the design of our probabilistic sensor model.
Third, our uncertainty-informed planner performs comparably to state-of-the-art open-vocabulary ObjectNav approaches that rely on a single fixed, hand-tuned prompt. We provide open-source code at: \url{https://github.com/PRBonn/uiap-ogn}.




\section{Related Work}
\label{sec:related}

Approaches for ObjectNav include end-to-end learning methods and modular methods. End-to-end learning methods~\cite{yadav2023iclr,ye2021auxiliary,li2023iros} map observations directly to actions, training visual representation and navigation policies simultaneously in simulation environments. While successful in simulation, these methods encounter challenges in real-world deployment due to sample inefficiency and their reliance on a limited set of closed-world vocabulary object categories. This limits their ability to semantically relate and detect novel objects, as well as adapt to unseen environments.

In contrast, modular ObjectNav methods~\cite{gadre2023cow,yokoyama2024vlfm,zhou2023icml,yu2023iros,chen2023rss,zhang2024iros} decompose the task into perception, mapping, exploration and point-goal navigation. Approaches build maps of the environment using geometric~\cite{stachniss2016handbook-slamchapter} and semantic information~\cite{endres2009rss}. A planner selects waypoints for exploration of the environment, while point-goal navigation determines the actions required to reach them. 

Recently, VLMs have significantly advanced robotic systems by enabling open-vocabulary object recognition and semantic reasoning. Early works such as VLMaps~\cite{huang23vlmaps} leverage VLM-derived latent features to build 2D geometric-semantic maps by projecting them into a 3D geometric representation. However, the feature fusion process in VLMaps is computationally expensive, making it unsuitable for real-time applications. Other approaches, including scene graph-based methods~\cite{hughes2022hydra,werby23hovsg}, construct hierarchical representations of semantic relationships and offer improved efficiency. Nevertheless, most scene graph representations remain too costly for real-time open-vocabulary exploration.

To enhance classical frontier-based exploration~\cite{yamauchi1997cira}, recent ObjectNav methods incorporate open-vocabulary perception using VLMs. Clip on Wheels (CoW)~\cite{gadre2023cow} was among the first to use CLIP-derived features~\cite{radford2021learning} for exploring frontiers until a target object is detected. Zhou et al.~\cite{zhou2023icml} use a CLIP-based object detector to detect objects and rooms and then use a language model to decide which room to navigate to. Recent works such as those by Chen et al.~\cite{chen2023rss} and VLFM~\cite{yokoyama2024vlfm} incorporate VLM-derived cosine similarity scores between the recorded camera image and a fixed text prompt. VLFM projects these scores onto a 2D grid to generate a semantic relevance map. This approach eliminates the need for expensive feature fusion by updating a single scalar value per cell at each timestep.

Despite these advances, existing open-vocabulary ObjectNav approaches typically treat VLM-based semantic relevance predictions as point estimates, ignoring the semantic uncertainty inherent in natural language and visual observations. ObjectNav methods, such as VLFM~\cite{yokoyama2024vlfm}, manually craft a fixed text prompt to elicit model responses, e.g. from BLIP-2~\cite{li2023blip2}, which result in strong ObjectNav performance. These approaches ignore the underlying semantic uncertainty of VLMs, although minor variations in prompt phrasing cause differences in semantic relevance, resulting in inconsistent navigation behaviours and performance. Notably, prompt engineering is hard to scale and brittle across environments. 



To address this, we propose an ObjectNav framework that explicitly models semantic uncertainty in open-vocabulary perception and integrates it into the planning process. Instead of relying on fixed prompts and treating similarity scores as point estimates, our method samples multiple prompt variants to approximate the distribution of similarity scores, thereby capturing semantic uncertainty using a new probabilistic sensor model.  We fuse these probabilistic semantic relevance scores into an online-built geometric-semantic map. Our frontier-based planner leverages this map to guide exploration, dynamically balancing semantic relevance and its associated uncertainty. In this way, our ObjectNav framework improves robustness to linguistic ambiguities and enables informed planning in complex environments.

\begin{figure*}[th]
  \centering
  \includegraphics[width=\textwidth]{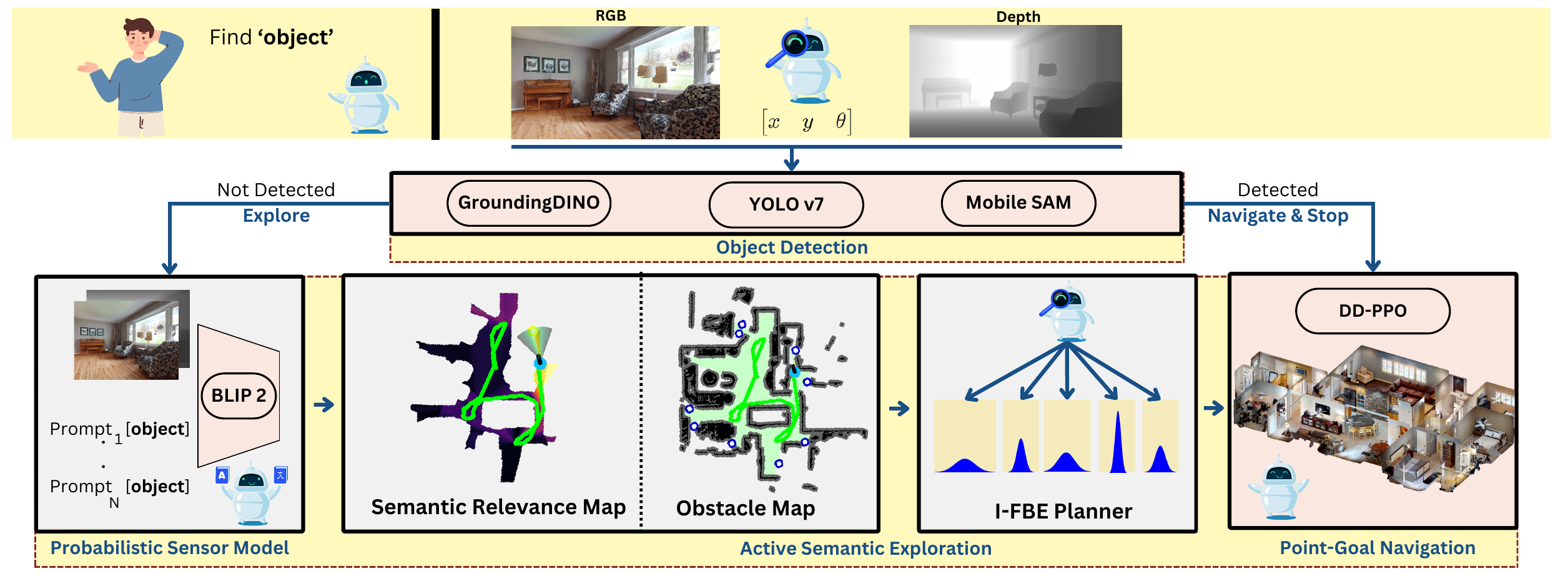}
  \caption{Our ObjectNav approach consists of the object detection and active semantic exploration modules. If the target object is detected in the currently recorded frame, the robot navigates directly to it using point-goal navigation, as discussed in~\secref{sec:object_detection}. Otherwise, it explores the environment using our uncertainty-informed frontier planner guided by our probabilistic semantic relevance map. Incorporating semantic cues and uncertainty into our map allows us to intelligently explore regions with higher probability of finding the target object.}
  \label{fig:pipeline}
\end{figure*} 

\section{Problem Formulation}
\label{sec:problem}

We formally define the ObjectNav task as an open-vocabulary, goal-directed navigation task~\cite{batra2020objectnav}. In ObjectNav, a robot is deployed in an unknown environment $E \subseteq \mathbb{R}^3$. The robot starts at an arbitrary pose $\mathbf{x}^w_0 = (\mathbf{v}^w_0, r^w_0)^\top$, where $\mathbf{v}^w_0 \in \mathbb{R}^3$ and $r^w_0 \in SO(2)$ are the robot's starting position and rotation in the world coordinate frame $w$. The robot's goal is to navigate to an object in the environment, e.g. a cup, referred to as the ``target object" $o \in \mathcal{O}$, where $\mathcal{O} \subset \mathbb{N}$ is a possibly infinite set of object categories present in the environment. The set of all target object instances in the environment is denoted as $\mathcal{G} = \{\mathbf{x}^{w} \in E \mid f(\mathbf{x}^w) = o\}$, where $f: E \to \mathcal{O}$ assigns an object category to each position in the environment. In each episode, the robot is tasked to navigate to a target object position $\mathbf{x}^w_g \in \mathcal{G}$. An episode $\tau = (E, o, \mathbf{x}^w_0)$ is defined by the a priori unknown environment $E$, user-defined target object category $o$, initial robot pose $\mathbf{x}^w_0$, and has a maximum length of $T \in \mathbb{N}$ timesteps. At each timestep $t$, the robot executes an action $a_t \in \mathcal{A}$, i.e. moving forward, turning left or right on the spot, or stopping. The robot receives an egocentric visual observation $\mathbf{I}_t = (\mathbf{I}^{\text{rgb}}_t, \mathbf{I}^{\text{depth}}_t)$ from a RGB-D camera, where $\mathbf{I}^{\text{rgb}}_t \in \mathbb{R}^{U \times V \times 3}$ is the RGB image and $\mathbf{I}^{\text{depth}}_t \in \mathbb{R}^{U \times V}$ is the depth image with resolution $U \times V$. The robot pose $\mathbf{x}^w_t$ at each timestep is assumed to be known. If the robot's distance $\lVert\mathbf{x}^w_t - \mathbf{x}^w_g\rVert_2$ to the target object position $\mathbf{x}^w_g \in \mathcal{G}$ is smaller than a clearance distance $c > 0$, $\mathbf{x}^w_g$ is visible from $\mathbf{x}^w_t$, and the robot stops, then the episode is successful. 

\section{Our Approach}
\label{sec:main}

Our uncertainty-informed ObjectNav approach explores the environment based on probabilistic estimates of semantic relevance. Our approach enables a robot to navigate to user-specified objects in an unknown 3D environment based on open-vocabulary perception, conceptually depicted in \figref{fig:pipeline}. We first perform object detection in each camera frame as described in \secref{sec:object_detection}. If the target object is detected, the robot directly navigates to it. Otherwise, we employ an active exploration strategy that identifies semantically relevant regions, such as a ``kitchen" for target objects like ``cup", by exploiting semantic correspondences and accounting for uncertainty using our probabilistic sensor model described in \secref{sec:sensor_model}. We construct a geometric-semantic grid map based on the new sensor model, updating occupancy and semantic relevance information online, as discussed in \secref{sec:mapping}. Our multi-arm bandit frontier-based planner uses our geometric-semantic map to guide the robot to regions with high semantic relevance, as outlined in \secref{sec:exploration}. 

\subsection{Object Detection and Point Goal Navigation}
\label{sec:object_detection}

We follow the approach of Yokohama et al.~\cite {yokoyama2024vlfm} for object detection. Each recorded RGB image is processed by the object detection module, which attempts to detect the target object $o \in \mathcal{O}$ in the frame using two object detectors. We rely on YOLOv7~\cite{wang2023yolov7}, a supervised learning-based detector trained on a pre-defined closed set of $91$ object categories $\mathcal{O}$, including a wide range of outdoor and indoor scenes. To allow our approach to work with potentially infinite and a priori unknown user-defined object categories of interest, we further use Grounding DINO~\cite{liu2023grounding} as an open-vocabulary object detector capable of detecting a wide range of objects specified by arbitrary text prompts. If the target object $o$ is detected, we use MobileSAM~\cite{zhang2023faster} to segment $o$ in the RGB image. To estimate the goal pose $\mathbf{x}^w_g$, the segmentation mask is applied to the depth image and projected into 3D world coordinates using known camera parameters. We use the centroid of the projected depth information of $o$ to determine a target object position $\mathbf{x}^w_g \in \mathcal{G}$. Then, we use DD-PPO~\cite{wijmans2020ddppo} to navigate to $\mathbf{x}^w_g$ from the current pose of the robot $\mathbf{x}^w_t$.

\subsection{Probabilistic Semantic Relevance Sensor Model}
\label{sec:sensor_model}
To extract semantic uncertainty from VLMs, we introduce our probabilistic sensor model. In our active semantic exploration module, we use the BLIP2~\cite{li2023blip2} VLM for semantically guided exploration. VLMs project both visual and textual inputs into a shared latent embedding space. Semantic relationships are preserved within this space, such that contextually similar images $\mathbf{I}^{\text{rgb}}$ and text prompts $p$ are embedded as latent representations $\mathbf{l}^{\text{rgb}} \in \mathbb{R}^D$ and $\mathbf{l}^p \in \mathbb{R}^D$ respectively, oriented in similar directions. We leverage the commonly used cosine similarity metric to quantify the alignment between text and image embeddings, providing an estimate of semantic relevance between an image and text: 
\begin{eqnarray}
  \label{eq:cosine_similarity}
    S(\textbf{l}^{\text{rgb}}, \textbf{l}^{p}) = \frac{\textbf{l}^{\text{rgb}} \cdot \textbf{l}^{p}}{\lVert\textbf{l}^{\text{rgb}}\|_2 \|\textbf{l}^{p}\rVert_2}\,.
\end{eqnarray}

However, 
VLMs are often sensitive to slight prompt variations, leading to fluctuating semantic relevance as measured in \eqref{eq:cosine_similarity}. We model the VLM's data uncertainty about semantic relevance as a Gaussian-distributed random variable 
\begin{eqnarray}
\label{eq:cosine_normal_distribution}
\mathcal{S}&\sim& \mathcal{N}(\mu_Z, \sigma^2_Z) \,,
\end{eqnarray}
where $\mu_Z$ is the mean semantic relevance score and $\sigma^2_Z$ its variance.
We approximate the mean and variance of $\mathcal{S}$ in a Monte Carlo fashion
by designing a prompt ensemble~\cite{lester2021promptensemble} using a set of $N$ prompts $\mathcal{P} = \{p_1, p_2, \dots, p_N\}$. We generate these prompts using the GPT4~\cite{openai2024gpt4technicalreport} language model, asking for alternative prompt formulations to the default prompts ``there is a \texttt{target\_object} ahead" and ``A \texttt{target\_object} is in the vicinity". The RGB image $\mathbf{I}^{\text{rgb}}$ and each prompt $p_i \in \mathcal{P}$ are mapped to their latent representations $\mathbf{l}^{\text{rgb}}$ and $\mathbf{l}^{p_i}$ by the VLM. Then, the mean $\mu_Z$ of the semantic relevance scores and their variance $\sigma^2_Z$ are:

\begin{align}
  \mu_Z & = \frac{1}{\vert\mathcal{P}\vert} \sum_{p_i \in \mathcal{P}} S(\mathbf{l}^{\text{rgb}}, \mathbf{l}^{p_i})\,,   \label{eq:semantic_relevance_mean}\\
  \sigma^2_Z & = \frac{1}{\vert\mathcal{P}\vert} \sum_{p_i \in \mathcal{P}} \big(S(\mathbf{l}^{\text{rgb}}, \textbf{l}^{p_i}) - \mu_Z\big)^2\,. \label{eq:semantic_relevance_variance}
\end{align}


Usually, not all parts of the sensor's field of view (FOV) contribute equally to semantic relevance. VLMs tend to assess semantic relevance more reliably for objects near the image centre, 
while peripheral regions often exhibit 
diminished influence on the semantic relevance. 
Following Yokohama et al.~\cite{yokoyama2024vlfm}, we address this limitation by introducing a viewpoint-dependent confidence measure $C_V$. However, unlike their approach of using the confidence measure in the map update, we use it in our sensor model. For image pixels close to the optical axis, $C_V$ is high, while it decreases towards the peripheral regions of the image:
\begin{eqnarray}
  \label{eq:variance_uncert}
  C_V(\theta) = \cos^2 \left(\frac{2\theta\pi}{\theta_{\text{fov}}}\right)\,,
\end{eqnarray}
where $\theta$ is the angle between the pixel and the optical axis, and $\theta_{\text{fov}}$ is the horizontal FOV of the robot's camera.

We combine this confidence measure with our image-based semantic relevance variance $\sigma^2_Z$ in \eqref{eq:semantic_relevance_variance} into a per-pixel semantic relevance variance $\sigma^2_Z(\theta)$: 
\begin{eqnarray} 
\label{eq:semantic_relevance_variance_total}
\sigma^2_Z(\theta) = \sigma^2_Z + \left(1 - C_V(\theta)\right) \,.
\end{eqnarray}

We use our new sensor model for semantic relevance in ObjectNav to update our probabilistic geometric-semantic map as detailed in the following \secref{sec:mapping}.




\subsection{Probabilistic Geometric-Semantic Mapping}
\label{sec:mapping}
Our geometric-semantic mapping stores and updates the belief about the initially unknown environment $E$ based on new incoming sensor observations. We maintain two grid maps, updating the geometric and semantic environment information received through depth sensor observations and semantic relevance estimation as described in \secref{sec:sensor_model}. The environment is discretised into a 2D top-down grid $G$ of resolution $H \times W$. At each timestep $t$, the geometric obstacle map $\mathcal{M}_{O,t} : G \to \{0,1\}^{H \times W}$ is updated based on the depth image's $\mathbf{I}^{\text{depth}}_t$ focal cone projected onto the 2D plane using occupancy mapping~\cite{yokoyama2024vlfm,moravec1985occupancy}. 

We update our novel semantic relevance map \mbox{$\mathcal{M}_{S,t}: G \to [0,1]^{H \times W}$} using our semantic relevance sensor model in \eqref{eq:cosine_normal_distribution}. 
We assume the prior belief $\mathcal{M}_{S,0}(m) \sim \mathcal{N}\left(\mu_{S,0}(m), \sigma_{S,0}(m)\right)$ of a grid cell $m \in G$ to be normally distributed with an uninformed prior mean $\mu_{S,0} = 0.5$ and large variance $\sigma_{S, 0}^2 = 0.5$, expressing that the environment is initially unknown. This choice of map prior enables us to formulate the posterior semantic relevance map update at timestep $t$ in closed form as the prior is conjugate, given the normally distributed sensor model $\mathcal{S}$. At timestep $t$, we update the map~$\mathcal{M}_{S,t}(m)$ for each grid cell in the camera's projected 2D focal cone:
\begin{align}
\mu_{S,t}(m) &= \frac{\sigma^2_{S,t-1}(m) \, \mu_{Z,t} + \sigma^2_{Z,t}(\theta)\, \mu_{S,t-1}(m)}{\sigma^2_{S,t-1}(m) + \sigma^2_{Z,t}(\theta)} \,, \label{eq:gauss_mean_update} \\
\sigma^2_{S,t}(m) &= \frac{\sigma^2_{S,t-1}(m)\, \sigma^2_{Z,t}(\theta)}{\sigma^2_{S,t-1} + \sigma^2_{Z,t}(\theta)} \,, \label{eq:gauss_uncertainty_update}
\end{align}
where $\mathcal{S}_t \sim \mathcal{N}(\mu_{Z,t}, \sigma_{Z,t}^2(\theta))$ is the current semantic relevance measurement computed as in \eqref{eq:semantic_relevance_mean} and \eqref{eq:semantic_relevance_variance_total}. As in the geometric obstacle map update, the semantic relevance measurement $\mathcal{S}_t$ is projected to the 2D plane using the camera's focal cone. Hence, we use $\mu_{Z,t}$ to update all grid cells $m$ in the focal cone while $\sigma_{Z,t}^2(\theta)$ is the variance of a pixel described by $\theta$ that is projected onto grid cell~$m$.
In the following \secref{sec:exploration}, we detail how we leverage the geometric-semantic map in our new ObjectNav planning method to explore the environment towards the target object of interest in a semantically-targeted fashion.




\subsection{Uncertainty-Informed Exploration}
\label{sec:exploration}
Building on the probabilistic geometric-semantic map described in \secref{sec:mapping}, we propose a new uncertainty-informed planner to perform semantically-targeted exploration towards the target object of interest. Our planner frames frontier exploration~\cite{yamauchi1997cira} as a multi-armed bandit problem. The planner seeks to find the next-best frontier to navigate towards by balancing the exploitation of frontiers with known high semantic relevance with exploration of semantically uncertain frontiers. We treat available frontiers as arms, evaluating their potential utility towards finding the target object by employing decision-theoretic reward functions.

At each timestep $t$, we leverage our geometric obstacle map $\mathcal{M}_{G,t}$ to navigate to frontiers of known free space and unexplored space. Let $m_{i} \in G$ be a grid cell representing the center of a frontier $i \in \{1, 2, \ldots, F_t\}$, where $F_t$ is the number of frontiers in $\mathcal{M}_{G,t}$. We evaluate our current semantic relevance map belief $\mathcal{M}_{S,t}(m_i) \sim \mathcal{N}(\mu_{S,t}(m_i), \sigma_{S,t}^2(m_i))$ at each frontier $i$. To guide the robot's exploration, we develop two planners, called I-FBE1 and I-FBE2. 

\textbf{I-FBE1} uses the expected improvement function~\cite{jones1998expimp} to select a frontier to explore among all available frontiers based on their semantic relevance and uncertainty, as:

\begin{align}
\label{eq:expected_improvement}
\mathrm{EI}(m_i) =\ &\left(\mu_{S,t}(m_i) - \mu^*_t\right)\Phi\left(\frac{\mu_{S,t}(m_i) - \mu^*_t}{\sigma_{S,t}(m_i)}\right) \nonumber \\
&+  \sigma_{S,t}(m_i) \, \phi\left(\frac{\mu_{S,t}(m_i) - \mu^*_t}{\sigma_{S,t}(m_i)}\right)\,, \\ 
\mu^*_t =\ &\max_{j \in \{1, \ldots, F_t\}} \mu_{S,t}(m_j) \,,
\end{align}
where $\mu^*_t$ is currently the highest semantic relevance among all frontiers in $\mathcal{M}_{G,t}$, and $\Phi$ and $\phi$ are the standard normal distribution's cumulative distribution and probability density function. The term $\Phi(\cdot)$ reflects the probability that the candidate frontier will improve upon the current best, supporting exploitation, while $\phi(\cdot)$ promotes exploration of uncertain frontiers that may lead to high potential improvement.

\textbf{I-FBE2} uses the Gaussian process upper confidence bound (GP-UCB)~\cite{srinivas2012gpucb} reward function. GP-UCB, in contrast with expected improvement, promotes aggressive exploration of frontiers, even if they are not promising for immediate gains:
\begin{eqnarray}
  \label{eq:gp_ucb}
  \text{GP-UCB}(m^i_t) = \mu_{S,t}(m^i_t) + \sqrt{\beta} \sigma_{S,t}(m^i_t) \,.
\end{eqnarray}

The first term $\mu_{S,t}(m^i_t)$ promotes exploitation by favoring regions with high predicted relevance, while the second term $\sqrt{\beta} \, \sigma_{S,t}(m^i_t)$ encourages exploration in areas of high uncertainty. The hyperparameter $\beta$ controls the trade-off. 




\section{Experimental Results}
\label{sec:exp}

%

%
We design our experiments to show the capabilities of our method. The results of our experiments support our key claims: (i) The success of prior open-vocabulary ObjectNav approaches relies on extensive prompt engineering, and prompt choice directly influences ObjectNav performance; (ii) We show that semantic relevance scores from a VLM prompt-ensemble are approximately normally distributed to validate the design of our probabilistic sensor model;
(iii)~Our uncertainty-informed planner performs comparably to the state-of-the-art open-vocabulary ObjectNav approaches that rely on fixed hand-engineered prompts.

\subsection{Experimental Setup}
\label{sec:setup}
\textbf{Datasets and hardware.} To assess the effectiveness of our ObjectNav approach, we conduct experiments in realistic indoor environments using the Habitat simulator~\cite{savva2019habitat}, a popular simulator for ObjectNav evaluations~\cite{batra2020objectnav}. We use two datasets, Matterport3D~(MP3D)~\cite{chang2017mp3d} and Habitat-Matterport 3D~(HM3D)~\cite{ramakrishnan2021hm3d}, both offering complex 3D reconstructions of indoor spaces. MP3D provides 3D scans with rich semantic annotations but limited unique environments and data imperfections. HM3D offers 1,000 high-quality scans with complex layouts and varied conditions. We evaluate our approach across varied, realistic scenes in 2,195 MP3D and 2,000 HM3D episodes, which contain 20 scenes, six object categories and 11 scenes and 21 object categories, respectively, following prior work~\cite{yokoyama2024vlfm}. Experiments are conducted on a workstation with a 12th-generation Intel Core i7 CPU, 64 GB RAM, and an NVIDIA RTX A5000 GPU.

\textbf{Evaluation metrics.} We evaluate each approach using Success Rate (SR) and Success Weighted by Path Length (SPL)~\cite{batra2020objectnav}. SPL measures the efficiency of the robot’s path by comparing it to the shortest possible route from the starting point to the nearest instance of the target object in the ground truth. If the agent fails to reach the target, the SPL score is zero; otherwise, it represents the ratio of the shortest path length to the agent’s actual path length, with higher values indicating better path efficiency. 

\textbf{Baselines.} To benchmark our planners I-FBE1 and I-FBE2, we compare our pipeline with geometric variants of the frontier planner~\cite{yamauchi1997cira}: Closest-FBE and RandomFBE. Closest-FBE directs the robot to the nearest frontier for exploration, while Random-FBE selects frontiers at random. We also evaluate against open-vocabulary semantically informed approaches, VLFM~\cite{yokoyama2024vlfm}, CoW~\cite{gadre2023cow}, ESC~\cite{zhou2023icml}, and ZSON~\cite{majumdar2022zson}. ZSON projects the image of the target object and the prompt into the same embedding space using the CLIP~\cite{radford2021learning} VLM and performs ObjectNav using a trained planning network. CoW uses gradient-based visualisation of the CLIP VLM along with frontier-based exploration, whereas ESC uses a VLM with a language model to guide frontier exploration. VLFM is an open-vocabulary frontier planner that uses VLM-derived cosine similarities to identify the most relevant frontiers. VLFM does not account for the uncertainty inherent in the VLM predictions and relies on a fixed prompt ``Seems like there is a \texttt{target\_object} ahead", making it a relevant baseline for evaluating the benefits of explicitly modeling semantic uncertainty.

\begin{table}[t] \footnotesize
\centering
\resizebox{\columnwidth}{!}{%
\begin{tabular}{lcc}
\toprule
\textbf{Prompt} & \textbf{SR ↑} & \textbf{SPL ↑}  \\
\midrule
Seems like there is a \texttt{target\_object} ahead & 52.60 & 30.42  \\
\midrule
A place where \texttt{target\_object} can be found  & 51.00 & 29.71  \\
A \texttt{target\_object} can be in the vicinity    & 53.20 & 31.20  \\
Seems like a \texttt{target\_object} is ahead       & 53.20 & 30.50 \\
A \texttt{target\_object} is in the vicinity        & 51.65 & 28.67  \\
\texttt{target\_object} likely ahead                & 52.45 & 29.86  \\
\texttt{target\_object }                            & 50.60 & 28.28  \\
\midrule
Ours (I-FBE1)& 52.25 & 28.96 \\
Ours (I-FBE2)& 53.50 & 27.31  \\ 
\bottomrule
\end{tabular}%
}
\caption[Impact of prompt phrasing variations ]{Impact of prompt phrasing variations on success rate (SR) and SPL in downstream navigation tasks. The prompt formulation significantly influences the performance on the ObjectNav task on the HM3D Dataset. Our uncertainty-informed methods perform comparably to the default prompt in VLFM~\cite{yokoyama2024vlfm} (top).}
\label{tab:prompt_engineering}
\end{table}





\subsection{Effect of Prompt Phrasing on ObjectNav Performance}
Our first experiment is designed to demonstrate that the success of prior open-vocabulary ObjectNav approaches heavily depends on prompt engineering. In particular, we show that the choice of prompt directly impacts ObjectNav performance. We evaluate VLFM~\cite{yokoyama2024vlfm}, which uses a fixed prompt: ``Seems like there is a \texttt{target\_object} ahead". We provide this prompt to the GPT4~\cite{openai2024gpt4technicalreport} model and ask it to generate alternative formulations for the same target object, out of which we then use seven prompts to run ObjectNav evaluations. To assess performance, we only modify the prompt for the BLIP2 VLM in VLFM and evaluate on 2000 episodes of HM3D using the SR and SPL metrics.

Our results are presented in~\tabref{tab:prompt_engineering}. The choice of prompt for the VLM to estimate semantic relevance significantly impacts the overall effectiveness of the ObjectNav pipeline. For instance, a slight modification—changing the prompt to ``Seems like a \texttt{target\_object} is ahead" improved SR by 0.6\%, while using only the object name reduced it by nearly 2\% compared to the default hand-engineered prompt. These results highlight that minor prompt adjustments can lead to a difference in capturing semantic relevance, potentially adversely affecting ObjectNav performance. In contrast, our uncertainty-informed approaches I-FBE1 and I-FBE2 leverage all seven prompts to estimate semantic relevance. Particularly, I-FBE2 achieves performance close to the hand-crafted prompt in VLFM without manual prompt tuning.

\subsection{Prompt Ensembling for Semantic Relevance}
\begin{figure}[t]
  \centering
  \includegraphics[width=0.95\columnwidth]{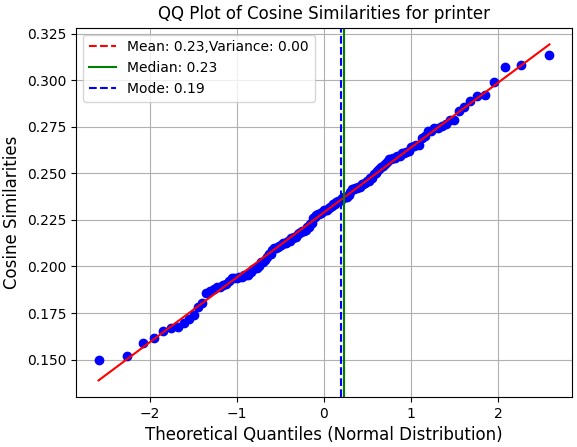}
  \caption[Quantile-Quantile plot of cosine similarities]{We display the Quantile-Quantile (QQ) plot of 100 VLM-predicted semantic relevance scores, which are generated from 100 unique prompts around the target object name ``printer". The quantiles of the semantic relevance scores~(vertical) are contrasted with the theoretical quantiles of the standard normal distribution~(horizontal). The linear relationship suggests that the semantic relevance scores are approximately normally distributed.}
\label{fig:qqplot}
\end{figure}
Our second experiment demonstrates that semantic relevance scores obtained from VLMs follow approximately a Gaussian distribution. This experiment validates the design of our probabilistic sensor model and the subsequent updates to its semantic relevance map. We use BLIP-2~\cite{li2023blip2} across five images and multiple object categories, including printer, oven, fridge, and table, with the target object category present in some images and absent in others. We generate prompt ensembles ranging from 10 to 100 variations for each category using GPT-4~\cite{openai2024gpt4technicalreport}. We then compute the cosine similarity between each prompt and image as in \eqref{eq:cosine_similarity}. 

We visualise the distribution of semantic relevance scores for the printer category in \figref{fig:qqplot} as a representative example. The results show that cosine similarities of prompt ensembles approximate a Gaussian distribution, as illustrated by the Quantile-Quantile (QQ) plot. These observations validate the design and assumptions of our sensor model in~\secref{sec:sensor_model}.




\begin{table}[t] \tiny
    \centering
    \resizebox{\columnwidth}{!}{%
    \begin{tabular}{lcccc}
        \toprule
        \textbf{Approach} & \multicolumn{2}{c}{\textbf{HM3D}} & \multicolumn{2}{c}{\textbf{MP3D}} \\
         \cmidrule(lr){2-3} \cmidrule(lr){4-5}
        & \textbf{SR}$\uparrow$ & \textbf{SPL}$\uparrow$ & \textbf{SR}$\uparrow$ & \textbf{SPL}$\uparrow$ \\
        \midrule
        Closest-FBE & 11.80 & 9.34 & - & - \\
        Random-FBE & 37.30 & 23.32 & - & - \\
        ZSON & 25.50 & 12.60 & 15.30 & 4.80 \\
        CoW  & - & - & 7.40 & 3.70  \\
        ESC  & 39.20 & 22.30 & 28.70 & 14.20  \\
        VLFM & 52.60 & \textbf{30.40} & \textbf{36.40} & \textbf{17.50}  \\
        \midrule
        Ours (I-FBE1)& 52.25 & 28.96 & 35.26 & 16.47 \\
        Ours (I-FBE2)& \textbf{53.50} & 27.31 & 35.63 & 16.52 \\ 
        \bottomrule
    \end{tabular}%
    }
    \caption[Benchmarking with state-of-the-art ObjectNav approaches]{We compare our approach with the baselines Closest-FBE, Random-FBE, VLFM and other recent approaches for ObjectNav on the HM3D and MP3D datasets. Our I-FBE1 and I-FBE2 perform comparably to the state-of-the-art ObjectNav methods.}
    \label{tab:baseline_comparison}
\end{table}

\subsection{ObjectNav Performance}

Our third experiment is designed to show that our uncertainty-informed planner performs comparably to state-of-the-art open-vocabulary ObjectNav approaches that rely on fixed hand-engineered prompts. We evaluate the two variants of our uncertainty-informed planner for the ObjectNav task across both MP3D and HM3D datasets and compare them to state-of-the-art approaches described in \secref{sec:setup}.

Our results are summarized in~\tabref{tab:baseline_comparison}. Frontier-based Closest-FBE planning performs worst, often trapping the robot in narrow, nearby regions. Frontier-based Random-FBE avoids this by selecting random frontiers, enabling more thorough exploration of unknown space. However, neither approach is semantically informed but purely geometric, resulting in suboptimal performance. Among the semantically-informed methods, in line with prior works, VLFM outperforms CoW, ESC, and ZSON, due to the strength of its underlying VLM and updating the map with cosine similarity values for each received camera frame. Our uncertainty-informed planners achieve success rates comparable to VLFM, despite not relying on hand-crafted prompts. Particularly, I-FBE2 outperforms VLFM across many prompt variations on HM3D as shown in~\tabref{tab:prompt_engineering}.  We demonstrate that open-vocabulary ObjectNav methods relying on a single hand-tuned prompt are brittle. In contrast, our uncertainty-informed planners leveraging our probabilistic sensor model offer a more reliable and robust alternative. However, our method consistently underperforms slightly on the SPL metric compared to VLFM. This is primarily because our planner is designed to actively explore regions with high semantic relevance uncertainty. As a result, the agent occasionally replans toward uncertain regions even if they are not highly semantically relevant, leading to detours en route to the target object. On our evaluation setup, each replanning step takes approximately 420 ms of wall-clock time per timestep.
\section{Conclusions and Future Work}
\label{sec:conclusion}
In this work, we proposed a training-free, open-vocabulary, semantic uncertainty-informed active perception pipeline for the ObjectNav problem in mobile robotics. We introduce a novel probabilistic sensor model based on prompt ensembles to estimate semantic relevance and uncertainty from VLMs. Our framework incorporates a probabilistic geometric-semantic map and uncertainty-informed frontier planners to address the brittleness of existing ObjectNav approaches, which rely on a single, fixed, hand-engineered prompt.
Our experimental results show that the success of prior open-vocabulary ObjectNav relies on extensive prompt engineering. 
In contrast, our uncertainty-informed planner performs comparably to state-of-the-art open-vocabulary ObjectNav approaches that rely on fixed hand-engineered prompts while reducing dependence on prompt engineering. Future work will investigate the real-world deployment of our approach in larger environments under sensor noise.

\bibliographystyle{plain_abbrv}

\bibliography{glorified,new}

\end{document}

%% file: stachnisslab-latex.tex
\usepackage{graphics}           
\usepackage{times}              
\usepackage{amsmath}            
\usepackage{amssymb}            
\usepackage{graphicx}
\usepackage{algorithm}
\usepackage[noend]{algpseudocode}
\usepackage{booktabs}
\usepackage{color}
\usepackage{listings}
\usepackage{subfiles}
\usepackage{hyperref}
\usepackage[nocompress]{cite} 
\definecolor{instructioncolor}{rgb}{.5,.5,.5}

\usepackage[font=small]{caption}

\def\secref#1{Sec.~\ref{#1}}
\def\figref#1{Fig.~\ref{#1}}
\def\tabref#1{Tab.~\ref{#1}}
\def\eqref#1{Eq.~(\ref{#1})}

\makeatletter
\usepackage{xspace}
\DeclareRobustCommand\onedot{\futurelet\@let@token\@onedot}
\def\@onedot{\ifx\@let@token.\else.\null\fi\xspace}

\makeatother

\usepackage{array}
\newcolumntype{L}[1]{>{\raggedright\let\newline\\\arraybackslash\hspace{0pt}}m{#1}}
\newcolumntype{C}[1]{>{\centering\let\newline\\\arraybackslash\hspace{0pt}}m{#1}}
\newcolumntype{R}[1]{>{\raggedleft\let\newline\\\arraybackslash\hspace{0pt}}m{#1}}